# Unified Emulation-Simulation Training Environment for Autonomous Cyber Agents

Li Li[1], Jean-Pierre S. El Rami[1], Adrian Taylor[1], James Hailing Rao[2], and Thomas Kunz[3]

[1] Defence Research and Development Canada, Ottawa, Canada
[2] Queen's University, Kingston, Canada
[3] Carleton University, Ottawa, Canada
`li.li2@ecn.forces.gc.ca`

**Abstract.** Autonomous cyber agents may be developed by applying reinforcement and deep reinforcement learning (RL/DRL), where agents are trained in a representative environment. The training environment must simulate with high-fidelity the network Cyber Operations (CyOp) that the agent aims to explore. Given the complexity of network CyOps, a good simulator is difficult to achieve. This work presents a systematic solution to automatically generate a high-fidelity simulator in the Cyber Gym for Intelligent Learning (CyGIL). Through representation learning and continuous learning, CyGIL provides a unified CyOp training environment where an emulated CyGIL-E automatically generates a simulated CyGIL-S. The simulator generation is integrated with the agent training process to further reduce the required agent training time. The agent trained in CyGIL-S is transferrable directly to CyGIL-E showing full transferability to the emulated "real" network. Experimental results are presented to demonstrate the CyGIL training performance. Enabling offline RL, the CyGIL solution presents a promising direction towards sim-to-real for leveraging RL agents in real-world cyber networks.

## 1 Introduction

The modern world relies heavily on the correct operation of communication networks in general and the Internet in particular. Rogue actors regularly attempt to disrupt such networks, comprising confidentiality, availability, and/or the integrity of essential information. Organizations and governments are therefore interested in hardening their systems, both by learning about possible attack vectors, and deploying suitable defenses. This is typically carried out as contests between a "red team" attacking the network and a "blue team" defending it, assembling a "game" between the red and the blue teams.

A key challenge is the scalability of the expert red and blue teams. Even though cyber defense and attack emulation tools provide the red and blue agents tools to automate the IT workflow by enabling the staging, scripting commands and filling in the payloads, the sequential decision making at each action step still relies on the human cyber expert.

Deep Reinforcement Learning (DRL) enabled autonomous agents are envisioned to carry out network cyber operations (CyOps) with superior decision-making capabilities. A growing body of work is exploring DRL agents for use cases from autonomous cyber penetration tests to network red team and blue team exercises [1-6]. Similar to other real-world applications such as self-driving vehicles and autonomous robots, the DRL algorithms may train the agents to learn and optimize their Course of Actions (CoAs) for multi-stage operations in the complex and dynamic environment of cyber networks.

Developing applicable agents in real cyber networks requires first a DRL agent training environment that models the cyber network where the agent will operate. The CyOp agent training environment has started drawing interest in the community. A few environments have been reported, open-sourced, or even offered as public challenges for training CyOp agents, e.g., training a blue defense agent against various red adversary agents [7-13].

A DRL training environment is often a simulator of the real environment. A simulator of the network CyOp environment is particularly challenging, given the complexity of modern networks and CyOp actions. Capturing configurations of and interactions across numerous network components, the simulator is prone to be incomplete, as it needs to represent the host and network states which are in fact unknown at times, even to an expert. The state changes caused by actions are stochastic instead of being deterministic. Changes in network configurations and in the involved CyOp action also often bring about code redesign of the simulator.

To manage this problem, current CyOp simulators [7-8, 10-12] are built on abstractions of actions and simplification of states. The agent trained from such a simulator however cannot be transferred or deployed to the real network, as the actions learned and state data used in training deviate from reality. For example, the real red team may apply tools to select from more than 10 different network discovery actions, each relying on different techniques with their corresponding network configurations. The red agent on the other hand is trained in the simulator on one abstract action of "network discovery" [10-13]. As a result, the trained agent is not transferrable to a real network.

CyOps grounded in realistic cyber networks share the same problem with many real-world RL applications: a good simulator is essential but hard to build [14-15]. To attain high fidelity, real system data may be used. However, gathering data from the real cyber network is equally time-consuming. Although the RL training environment for achieving sim-to-real has been well investigated and advanced in other domains such as robotics, for example, through using environment images directly [16], the solutions are not applicable to cyber networks.

This work investigates approaches for building the CyOp training environment towards the goal of sim-to-real agent training and transfer. A Cyber Gym for Intelligent Learning (CyGIL) is presented, which is a unified deployment across both the real (or emulated) CyOp network, namely CyGIL-E, and its mirroring simulator, namely CyGIL-S. CyGIL-E runs on the real network or its emulated version over virtualized hardware of VMs (Virtual Machine). Actions in CyGIL-E use operational tools as they are used in the real network. CyGIL-S is auto-generated using CyGIL-E data to mirror the real environment. To our knowledge, this is the first CyOp training environment unified on both real (emulated) and simulated cyber networks with a complete cross-training and evaluation loop. The contributions include the following:

1. A cyber network RL environment that supports efficient agent training with high fidelity in a unified emulator (or real network) and simulator
2. Unsupervised auto-generation of CyGIL-S (the simulator) from the real (or emulated) network
3. A unified DRL training framework across CyGIL-E and CyGIL-S, demonstrating effective representation learning to reduce the time for data collection and for agent training

The rest of the paper is organized as follows. Section 2 presents the CyGIL system including both CyGIL-E (emulation-based) and CyGIL-S (simulation-based) training environment. Section 3 elaborates on the issues in generating CyGIL-S through experimental results. Section 4 presents the unified agent cross-training and evaluation solution. Section 5 draws concluding remarks.

## 2 Unified CyGIL-E and CyGIL-S

### 2.1 System Design

CyOps involve sequences of actions and their impact on network states over a short or long period. In the process, the attacker, referred to as the red agent, takes a sequence of actions to form and complete a cyber-kill chain [17] to break the confidentiality, integrity and availability of the network information and services. Meanwhile, the defender, referred to as the blue agent, must sustain the network mission objectives, throttling the kill chain, removing the red relics and recovering the compromised functionality. Red and blue agents use their respective operation tools to conduct the described CyOps.

The conceptual framework of the unified CyGIL is shown in Figure 1 (with some of the notations introduced more formally in Section 2.2). Both red and blue agents choose, using appropriate tools, specific actions that are applied to the training environment. The training environment knows about the game objectives and returns both observations (success or failure of the chosen action, information gained from successful execution of an action, etc.) as well as a reward. The reward reflects how well the action advances the agents' objectives.

An implementation of the unified CyGIl-E and CyGIL-S is illustrated in Figure 2. The CyGIL network can be either a real network or its emulation on virtualized hardware, encompassing network assets including the CyOp tools, and additionally actors such as users. Fig. 2 presents the implementation details of the mini CyGIL configuration for research, where the network is emulated on virtualized Mininet switches using the Open Network Operating System (ONOS) Software Defined Network (SDN) controller. Large networks are emulated using vSphere [18].

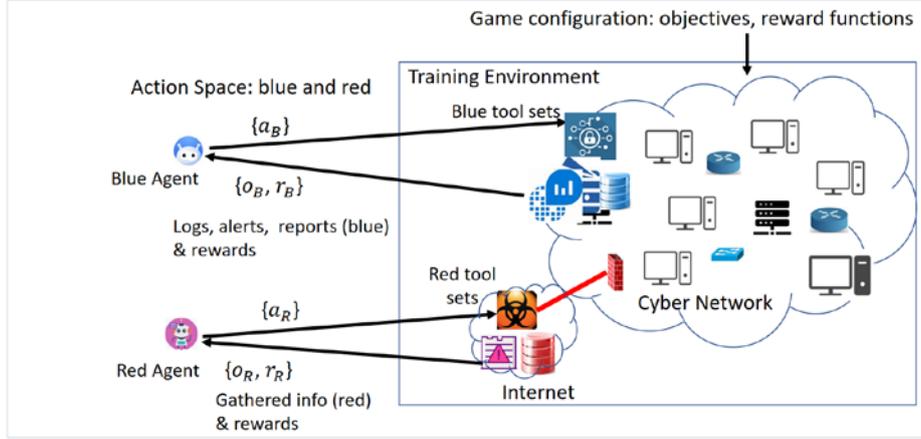

**Fig. 1.** Modeling CyOp environment to CyGIL framework: action $a_B \in A_B$, $a_R \in A_R$; observation $o_B \in O_B$, $o_R \in O_R$; reward $r_B$ and $r_R$ produced by $R_B$ and $R_R$ respectively

In the CyGIL training node, the CyGIL environment (env) library wraps the network and the training game into a CyGIL env python class to stand up the gym instance that provides the openAI gym interface [19] to the agent(s) for DRL training. Each gym training instance consists of the network and the training game. A training session may open its CyGIL gym instance in either CyGIL-E or CyGIL-S.

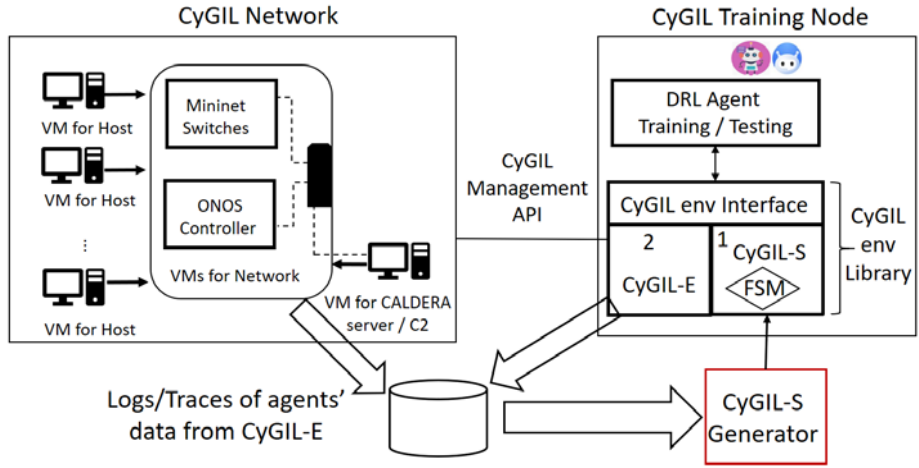

**Fig. 2.** CyGIL system implementation: 1 – agent training; 2 – representation learning, agent transfer and verification; dashed lines – the interface between CyGIL library and the real (or emulated) network and CyOp tools; C2: Command and Control of CyOp toolset(s)

In [9], it is demonstrated that red agents in CyGIL-E can learn and optimize their decision engines to achieve different attack objectives across the network. The agent learned, using the CALDERA red team tool [20] shown in Figure 2, to form an optimized end-to-end kill chain step-by-step, from network discovery, command-and-control, credential access, privilege escalation, defense evasion, lateral movement, to information collection and exfiltration, all starting from knowing nothing about either the network or the actions, i.e., what CALDERA can do. Training in CyGIL-E enables realistic and transferable agents, e.g., agents that attack networks using CALDERA, a SoTA tool used during human red team exercises to harden a network. Training the red agent in CyGIL-E requires from days to weeks, varied by the training games and algorithms used. The delay is mainly caused by the time taken for real action executions in the network, as well as resetting a network at the end of a training episode.

### 2.2 Unsupervised Auto-Generation of CyGIL-S

CyGIL-S is generated from the data collected in CyGIL-E (Figure 2) as follows. A CyGIL gym instance that consists of the network and the training game is modelled as a Markov Decision Process (MDP) $M = \langle S, A, P, R, s_0 \rangle$, where $S$ is the network state space and $A$ the action space. P is the probability defined on $S \times A \times S$, with the probability that action $a$ in state $s$ at time t will lead to state $s'$ at time $t+1$, written as $P_a(s, s') = \Pr(s_{t+1} = s' | s_t = s, a_t = a)$, $a \in A$, $s, s' \in S$. R is the reward function defined on $S \times A \times S \to \mathbb{R}$, with $R_a(s, s')$ as the reward function after transitioning from state $s$ to state $s'$, due to action $a$. The initial distribution of $S$ is $s_0$.

From $M$, $GN = \{S, A, P, s_0\}$ defines the training scenario that consists of only the network and the action spaces. On a $GN$, multiple training games can be designed, differentiated by their reward functions $R(s_t, a_t, s_{t+1})$ and the game ending criteria. The $GN$ embodies a Finite State Machine (FSM) from which the simulator of the training scenario can be built.

The state space $S$ is however unknown, except represented partially by measurements which may be more observable to the blue agent than the red agent. However, during the training, an agent only needs to see its own observation space. We thus approximate $GN^* \cong GN$ and decompose $GN^* = \{GN_1, GN_2, \ldots GN_M\}$ on a per-agent basis. In $GN_k$ for agent k, 1≤k≤M, the unknown $S$ is replaced by the agent's observation space, e.g., $GN_R = \{O_R, A_R, P_R, s_{0R}\}$ for the red agent.

$FSM_k$ is thus constructed for $GN_k$ without the unknown ground truth of $S$. Let the red agent carry out actions on $GN_R$ in CyGIL-E and gather tuples $(a, o, o') \in \mathcal{D}$. Assume action $a$ taken at the input $o$ leads to $N$ different outputs $o'_1, o'_2, \ldots, o'_N$. The transition probability $P$ is calculated as

$$P_a(o, o'_i) = \frac{C^{o'_i}_{(a,o)}}{\sum_{j=1}^N C^{o'_j}_{(a,o)}}, \quad \sum_{j=1}^N P_a(o, o'_j) = 1 \qquad (1)$$

where $P_a(o, o'_i) = \Pr(o_{t+1} = o'_i | o_t = o, a_t = a)$, and $C^{o'_i}_{(a,o)}$ counts the output observation $o'_i$ when action $a$ is taken at the input observation $o_t = o$. In the case where multiple agents take actions at each action step, $a$ extends to a vector to account for the different actions taken by agents. The details are not included here for brevity.

The CyGIL-S generated in this way produces observation data $o_i'$ upon receiving $(a, o)$, forming the transition $(a, o, o_i')$ according to $P_a(o, o_i')$ for agent training. This enables a lightweight CyGIL-S that supports fast agent training, as shown in the next section. The CyGIL-S generation is data-centric, agnostic to network topology, action sets, any game parameters and any values of the data. The CyGIL-S generator code is therefore reusable for different *M* and *GN* when the network and/or game changes.

## 3    CYGIL-S Evaluation

### 3.1    Test Data Collection Time

A key challenge is to collect, in as little time as possible, a data set *D* that can generate a sufficient CyGIL-S. A sufficient CyGIL-S embeds all required state space and transition probabilities to train the agent to the optimal policy. Data collection in the real-world physical system is time-consuming and expensive. Therefore, identifying the sufficient set *D* is critical.

An example is used here to illustrate the time expense issue. The experiment network is depicted in Figure 3. All hosts inside the network can reach the Active Directory Server and its Domain Controller (DC) on host 6 which is a Windows 2016 server. Other hosts in the network are Windows 10 machines except for hosts 1 and 9 which run on Linux Ubuntu 18. Hosts 1 and 2 are reachable from the external "Internet" where the red agent is trained to operate the attacker's C2. Hosts on the same switch belong to the same subnet and can communicate with each other. Between different subnets, firewall rules controlled by ONOS allow host 5 to communicate with hosts 2 and 3 in addition to hosts in its subnet. Each host communicates with some other hosts at any given time, forming the user traffic.

The training game defines the agent's objective as to land on the DC of ADS on host 6. If it succeeds, the red agent will have the admin privilege to breach the entire domain. The red agent's action space is shown in Figure 4, encompassing key tactic groups in the ATT&CK framework. These actions mimic the general tactics of several Advanced Persistent Threat (APT) groups [17] to enable the network end-to-end kill chain. As the initial state of the training game, the red agent has already compromised host 2 via phishing and implanted a "hand", i.e., malware. The hand reports back to the agent at C2, as supported by the CALDERA staging framework.

The reward function R for each action step is defined as $R = G - L$ where G is the gain and L is the cost. The game sets $G = 0$ for any action unless the agent reaches the objective, where it receives $G = 100$. It also sets $L = 1$ and $L = 8$ per hand, for an action that has all the required input parameters and an action that does not, respectively. This assigns a higher penalty to the selection of actions which do not even have the required parameters to form the execution commands. The reward after each action step is thus always negative unless the objective is reached. The game ends when either the agent achieves the objective or the game reaches the maximum number of steps which is set to 80. When failing to achieve the objective, the red agent ends up with a negative accumulated return of -80 or lower. The most optimized CoA can reach the objective in 6 steps, with some steps requiring concurrent actions taken by multiple hands. The red agents achieves an accumulated reward of 92 in this case, provided the random action outcomes all favor the agent.

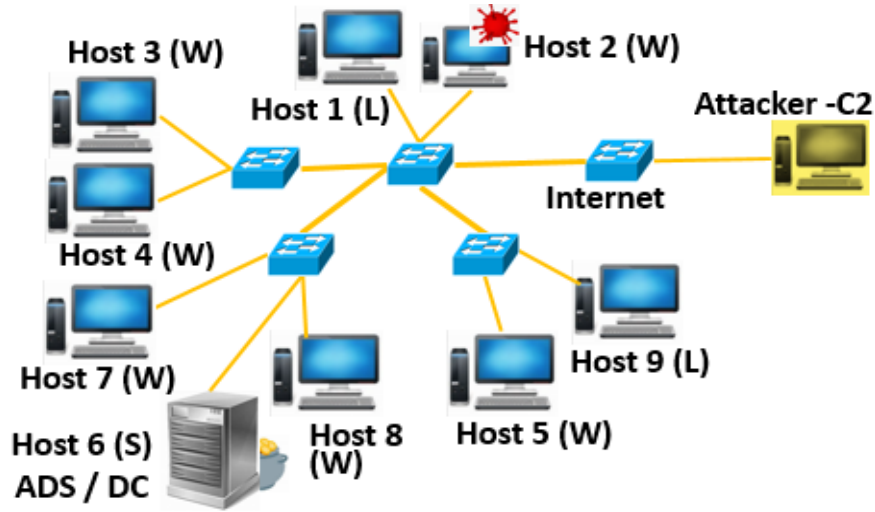

**Fig. 3.** The Example Network

| ATT&CK Tactics | ATT&CK Techniques | Game action ID |
| --- | --- | --- |
| Discovery | T1135 – Network Share Discovery | 0 |
| Discovery | T1087.002 - Enumerate AD user accounts | 11 |
| Discovery | T1018: Enumerate Active Directory computer objects | 12 |
| Discovery | T1016: collect ARP details | 13 |
| Discovery | T1003: reverse nslookup | 14 |
| Credential Access | T1003.001: Mimikatz to extract credentials | 7 |
| Credential Access | T1110.001: Brute Force credentials of domain user (NTLM or Kerberos) | 10 |
| Privilege Escalation | T1548.002: Download and execute Sandcat file agent as admin user | 8 |
| Lateral Movement | T1021.006: Execute Sandcat from fileshare remotely with WinRM | 1 |
| Lateral Movement | T1021.006: Execute Sandcat from fileshare remotely using PsExec | 2 |
| Lateral Movement | T1021.006: Copy Sandcat File with SCP and execute using PsExec | 3 |
| Lateral Movement | T1021.006 Execute Sandcat File remotely from system using WinRM | 4 |
| Lateral Movement | T1021.006 Mimikatz PSH and PsExec for launch Sandcat file on remote machine | 9 |
| Lateral Movement | T1021.006: PsExec to copy and launch Sandcat file on remote machine | 15 |
| Lateral Movement | T1570: Copy Sandcat file to remote system using WinRM and SCP | 6 |
| Lateral Movement | T1570: Copy Sandcat file to file share | 5 |

**Fig. 4.** Agent Action Space – key TTPs from ATT&CK framework

Emulating the network on a Windows laptop that runs on Intel(R) Core (TM) i9 and 64 GB RAM, the agent is trained using CyGIL-E on a second laptop that has an Intel(R) Core (TM) i7 and 64 GB RAM. The time to reach the optimized policy ranges from 7 to 20 *days*, depending on the learning algorithms used [9]. Using data collected from CyGIL-E to generate CyGIL-S according to Equation (1) and configuring the same game on CyGIL-S to train the agent afresh, the DQN [21], PPO [22] and C51 [3] Rainbow agent can all learn the optimal policy with average training times of 17.31, 25.92 and 5.76 *minutes* respectively, much faster than in CyGIL-E.

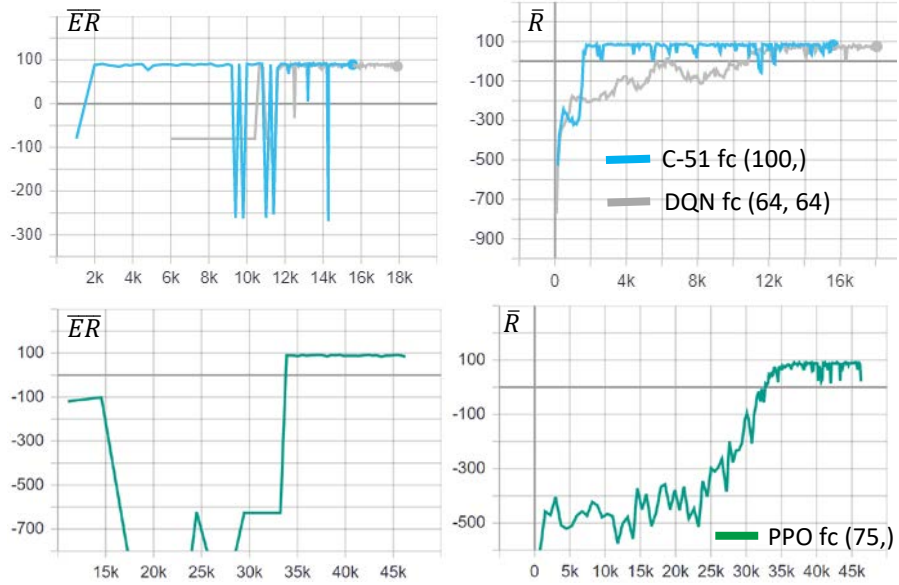

**Fig. 5.** Agent training in CyGIL-S - X axis: training steps; $\overline{R}$: Average training reward; $\overline{ER}$: Average evaluation rewards, fc: the architecture of the fully connected layer

Different data sets D from CyGIL-E are tested in generating CyGIL-S. From Table 1, to generate a sufficient CyGIL-S that can train the agent to the optimal policy, the time required approximates that for training the agent in CyGIL-E. This time is unsatisfactorily long. Although algorithm parameter tuning, tests of different algorithms and agent training with new reward functions and game–ending criteria can be carried out on CyGIL-S as shown in Figure 5, which are not feasible on CyGIL-E, reducing the data collection time is desirable. This is discussed in the next section.

**Table 1.** Data Set for Generating CyGIL-S

| Source of Data Set $D$ | Sufficient CyGIL-S Generated |
|---|---|
| Random Plays of 10 days | No |
| A DQN training session | Yes |
| A PPO training session | Yes |

### 3.2 Unknown Transitions in CyGIL-S

Both training and evaluation in CyGIL-S experience unknown states and action pairs. This is because the agent in CyGIL-S may step into the states that have not been reached in CyGIL-E, nor embedded in the data set *D* collected, given the large state space. The data-based simulated RL training environments in every real-world application face the recurring problem of too little data, given the cost of collecting data [14]. Thus unknown $P_a(o, o')$ for some transition are always encountered in CyGIL-S, even though it is a sufficient CyGIL-S for certain $(a, o)$ sets.

In CyGIL-S, a $(a, o)$ combination without a known $o'$ to complete the transition of $(a, o, o')$ is processed in CyGIL-S by setting $o' = o_I$, s.t. $P_a(o, o_I) > P_a(o, o')$, $\forall o' \neq o_I$. For red agents, $o_I = o$ often holds, because an action is most probably not executable. This is not the norm for blue agents, however. More data will support a better transition approximation for unknown input combinations.

The histogram of unknown transitions encountered in the 25 training sessions across two sufficient CyGIL-S instances which are generated from different DQN training session data in CyGIL-E, is illustrated in Figure 6. Even though these unknown state transitions in CyGIL-S do not seem to significantly impact the training results, we need to address them. The unified training solution described next leverages them to reduce both data collection and training latency.

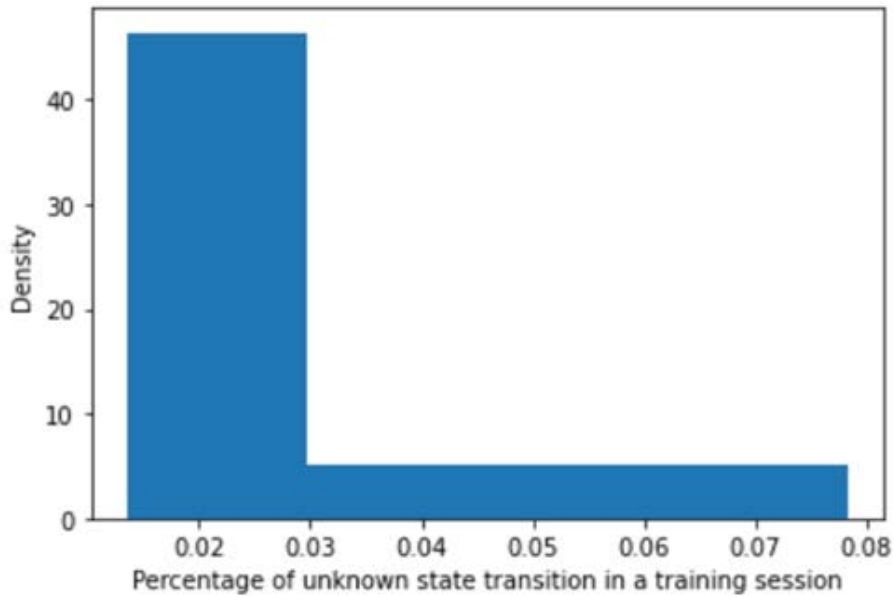

**Fig. 6.** Unknown transition distribution in sufficient CyGIL-S

## 4 Unified CyGIL Training

### 4.1 The Cross Training Loop

To reduce the required time for data collection in CyGIL-E and the overall agent training time, a unified CyGIL-E and CyGIL-S solution is developed. Its mechanism consists of a closed loop of transfer and continuous learning, as illustrated in Figure 7. The same experiment example (Figure 3 and Figure 4) is used to illustrate the details.

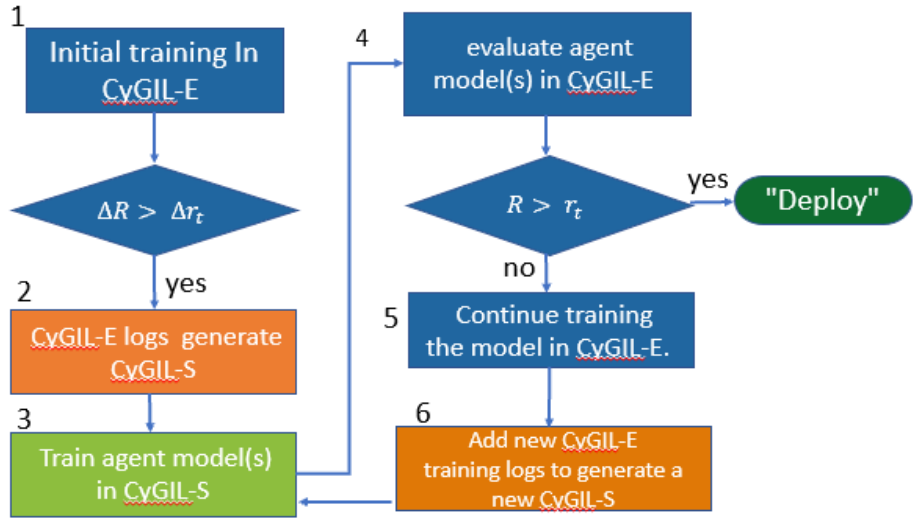

**Fig. 7.** Agent training in unified CyGIL-E and CyGIL-S

As shown in Figure 7, the unified training loop starts in Segment (SEG) 1 by training the agent in CyGIL-E until the reward improvement jumps over a threshold, $\Delta R > \Delta r_t$. At this stage, the training average reward is often far below the optimal value and the evaluation reward does not yet show any improvement. However, some good paths have already been traversed in the state space. This training session is used as a representation learning to move towards the better region(s) of the state space. Using the above network and game example, a sample efficient algorithm DQN is used in this SEG, as shown in Table 2. After 113 training episodes with more than 8k steps, the training average reward improved to -0.9 compared to the initial value of -912, with the episode length (the number of steps it takes the red agent to either successfully complete an episode or reach the end of an episode training) reducing from 80 to 24. The average performance of the subsequent training episodes in CyGIL-E degraded as shown in Figure 8 (a) and (b), as expected, as the model is far from being converged.

**Table 2.** Training Results across CyGIL-E & CyGIL-S

| Loop SEG & Model | In CyGIL-E or S | Elapsed Time | Training Average Reward and Best Episode Length |
|---|---|---|---|
| 1 - DQN | CyGIL-E | 35.5 h | -0.9, 24 |
| 3 - PPO | CyGIL-S | 29 m | 26.5, 10 |
| 5 - PPO | CyGIL-E | 1.1 h | 26.5, 10 |
| 3 - C51 | CyGIL-S | 4 m | 92, 8 |

The training loop moves to SEG 2, trigged by meeting the criteria for entering the SEG. CyGIL-S is generated from the data collected up to this point. The generation of CyGIL-S takes only a few seconds. Then the agent is trained in the CyGIL-S using PPO in SEG 3. Though it cannot be trained to converge to the optimal policy, the agent is much further improved through this fast training in CyGIL-S.

As shown in Table 2 and Figure 8 (c), in training in CyGIL-S, the model improves in its training reward and episode length and yet performs poorly in average evaluation reward in CyGIL-S. This CyGIL-S contains only limited state space. Many action and input observation pairs $(a, o)$ are not yet found in the CyGIL-S since they have not been encountered in CyGIL-E yet. That is, many unknown transitions may be encountered in the current CyGIL-S. As described in Section 3.2, unknown transitions cause a high penalty to R due to $L = 8$ per hand, because unknown transitions frequently result in non-executable actions. This then pushes the CyGIL-S training to explore and exploit states that are already embedded in the data from CyGIL-E.

The CyGIL-S training indeed wants to move fast in the region that has already been covered in data from CyGIL-E. While CyGIL-E has collected the data for a region, the better paths that can be achieved in this region are not yet reached in CyGIL-E, given the slow action execution in CyGIL-E. It should be noted many more paths are in the data from CyGIL-E than the paths that have been executed in CyGIL-E, because many new paths can be formed by concatenating the state transitions in the data. Stepping through paths is very slow in CyGIL-E. The CyGIL-S training compensates for the latency problem by quickly exploring and exploiting the potential paths to generate a better model.

When its policy cannot be improved further in the current CyGIL-S, the agent is transferred to CyGIL-E as shown in SEG 4. In CyGIL-E, if the agent model already exceeds the required return $(r_t)$, the agent training is completed and the agent can be deployed in the real (emulated) network. Otherwise, the agent model continues its training in CyGIL-E as shown in SEG 5, this time leveraging the knowledge it obtained from training in CyGIL-S. This quickly leads the agent to explore regions with much higher returns than during its previous training session in CyGIL-E.

Moving from SEG 5 to SEG 6, i.e., returning to continued training in CyGIL-S, is triggered by counting the number of training episodes CyGIL-E. After every 4 episodes, the newly collected data is added to the previously collected logs to generate a new CyGIL-S as shown in SEG 6. Using this new CyGIL-S, training loops back to SEG 3 for agent training in CyGIL-S while the training in CyGIL-E also continues in parallel to collect more data from further training episodes. Again, in CyGIL-S, the agent model is trained from scratch without using the model from CyGIL-E, given the fast training in simulation. For this training game scenario, after collecting data from an additional 8 episodes in CyGIL-E, the new CyGIL-S trains successfully an optimized agent policy using the C51 rainbow algorithm [19] (Table 2 and Figure 8 (d)). This trained model achieves the optimal CoA when transferred and evaluated in CyGIL-E, in all 50 evaluation runs.

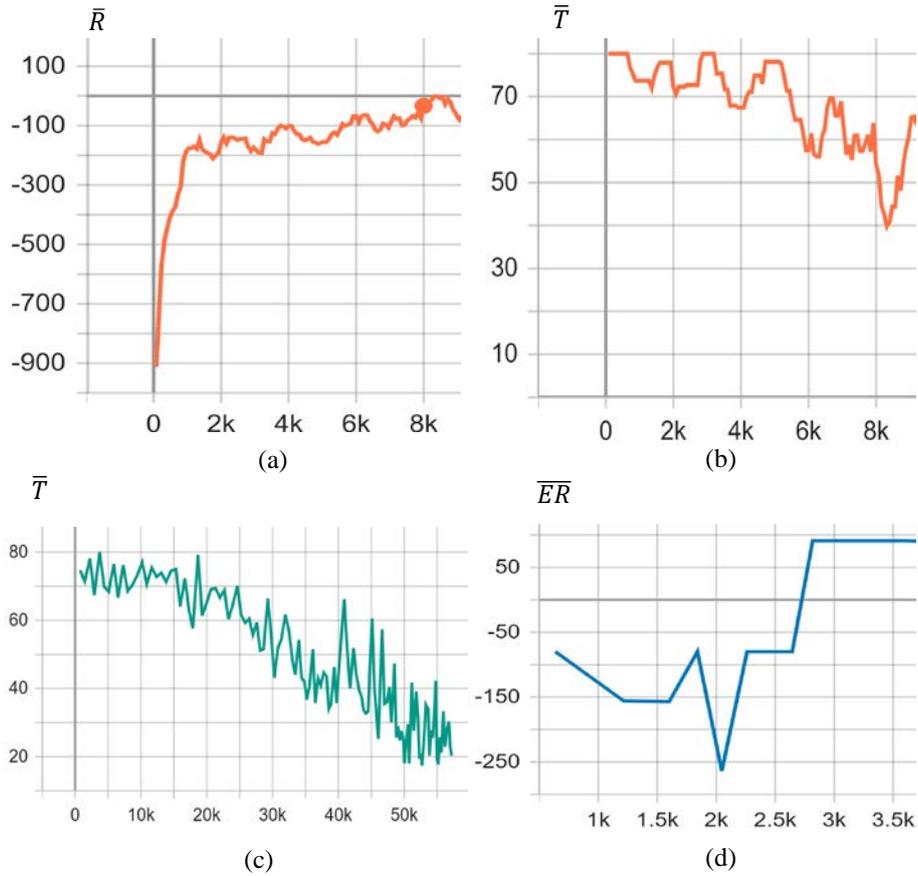

**Fig. 8.** Unified training across CyGIL-E and CyGIL-S - X axis: training steps; $\bar{R}$: Average training reward; $\bar{T}$: Average episode length; $\overline{ER}$: Average evaluation rewards; (a) and (b) Initial SEG 1 in CyGIL-E using DQN algorithm; (c) First training in CyGIL-S using PPO algorithm (SEG 3); (d) Second training in a new CyGIL-S using the C51 rainbow algorithm (SEG3)

It is noted that when the last 8 episodes were executed within CyGIL-E, the model training in CyGIL-E was still far from converging to the optimal policy. The average episode length in CyGIL-E already improved to around 10. However, the training still requires a long time before approaching the optimal policy. The reduced learning rate, the selected "good" actions which are always executable, and the added evaluation episodes all extend the time required in CyGIL-E. Yet, the collected data already embed the best CoA and enable CyGIL-S to train the agent to the optimal policy.

### 4.2 Discussions

The agent training in CyGIL-E can be considered a form of representation learning to find the right set of data to collect, which embodies the training path towards the optimal model. This is much more efficient than collecting random data from a real network or emulated network without training the agent. The CyGIL-S training rapidly pushes the agent model towards the better areas in the region, towards the optimal CoA. The model trained in CyGIL-S is transferred and continuously trained in CyGIL-E to collect further data in more promising areas of the state space, compared with the previous session. It also further advances toward the optimal CoA. This iterative process finally generates a sufficient CyGIL-S that can train the agent to reach the optimal CoA.

The sample efficient algorithm is preferred for the initial representation learning in CyGIL-E for good exploration. An on-policy algorithm that can converge fast is currently selected to advance the agent model in the CyGIL-S and further in CyGIL-E for efficiency. The C51 rainbow algorithm, which is an effective Categorical DQN algorithm, is found to train very well in this experiment.

In amulti-modal state space, the model trained in CyGIL-S may arrive at a poor local optimum. In current tests, the continuous training in CyGIL-E has been successful in stepping out and continuing towards the optimal area of the states. This may be due to the stochastic gradient descent and other techniques used in the DRL models which have the strong property in getting out of a poor local optimum. Though the unified training loop is implemented and found to be effective in the initial experiment scenarios, the segment transfer trigger points, the selection of the DRL algorithms in each training segment and the effectiveness in handling local optima across CyGIL-S and CyGIL-E are only heuristics at present. Further research is required to solidify these parameters across different CyOp scenarios.

## 5    Concluding Remarks

This paper presents our approach for building a CyOp training environment towards the goal of sim-to-real agent training and transfer. A Cyber Gym for Intelligent Learning (CyGIL) is presented, which is a unified deployment across both the real (or emulated) CyOp network, namely CyGIL-E, and its mirroring simulator, namely CyGIL-S. CyGIL-S is generated automatically from trace data collected in CyGIL-E and allows us to train autonomous agents with high fidelity: agents trained in CyGIL-S use the same action space that agents will encounter in real networks, and are therefore directly deployable, unlike other CyOp training environments based on simulation. Training agents in CyGIL-S takes only a fraction of the time it would take to train these agents in the real/emulated network and allows us to explore various "what-if" scenarios that would otherwise be infeasible.

A challenge in building CyGIL-S is to identify how much data needs to be collected from CyGIL-E to build a sufficient simulator (i.e., a simulator that allows a trained agent to discover the optimal course of action). In the initial version, as summarized in Table 1 at the end of Section 3.1, we would require CyGIL-E to complete the training of a single agent to collect enough data to build a sufficient CyGIL-S. While this may seem unattractive, once this CyGIL-S is constructed, we can then use it to explore additional training algorithms, modify game objectives, etc., with little additional runtime cost.

A key challenge in collecting fewer data is the occurrence of unknown state transitions: the FSM underlying CyGIL-S depends on having observed state (or observation) changes as a consequence of actions taken by an agent in the real or emulated network. For the sufficient CyGIL-S described in Section 3, only a small number of unknown state transitions are encountered during training and they do not prevent the trained agent from learning the optimal course of action. However, reducing the collected data will increase the number of these unobserved state transitions, so any reductions have to carefully manage this problem.

Our solution, described in Section 4, uses a unified training approach combining both CyGIL-E and CyGIL-S. In a nutshell, we collect initial data to build a first CyGIL-S. An agent trained with this incomplete simulator will move relatively quickly towards more promising courses of actions. Transferring the agent back into CyGIL-E, we can collect more relevant state transitions as the more knowledgeable agent explores more promising regions of the state space. With this collected data, we can then build better versions of CyGIL-S, and in our running example, the trained agent learns the optimal course of action after one iteration through the loop. Overall, as summarized in Table 2, this speeds up agent training compared to training an agent purely in CyGIL-E. It also significantly reduces the time to build a sufficient CyGIL-S.

The approach presented in this paper seems promising. Future work will expand on the key iterative loop shown in Figure 7: what parameters will trigger switches between CyGIL-E and CyGIL-S, what training algorithms to best use in different stages of the training cycle, etc. We will also explore additional scenarios to explore in more depth how robust this approach is against getting stuck in local optima.

## References

1. S. Chaudhary, A. O'Brien, and S. Xu, "Automated post-breach penetration test-ing through reinforcement learning", in *Proceedings of 2020 IEEE Conference on Communications and Network Security (CNS)*, 2020.
2. M. C. Ghanem and T. M. Chen, "Reinforcement Learning for Intelligent Penetration Testing", in *Proceedings of Second World Conference on Smart Trends in Systems, Security and Sustainability* (WorldS4), pp. 185–192. 2018
3. H. Nguyen, H. N. Nguyen, and T. Uehara, "Multiple Level Action Embedding for Penetration Testing," The 4th International Conference on Future Networks and Distributed Systems (ICFNDS), 2020

4. F. M. Zennaro and L. Erdodi, "Modeling penetration testing with reinforcement learning using capture-the-flag challenges and tabular Q-learning," *arXiv preprint arXiv:2005.12632*, 2020
5. J. Schwartz and H. Kurniawati, "Autonomous Penetration Testing using Reinforcement Learning," *CoRR, vol. abs/1905.05965*, 2019
6. M. Sutana, A. Taylor and L. LI, "Autonomous network cyber offence strategy through deep reinforcement learning", in *Proceedings of SPIE conference on Defences and Commercial Sensing*, 2021, April 2021
7. C. Baillie, M. Standen, J. Schwartz, M. Docking, D. Bowman and J. Kim, "CybORG: An Autonomous Cyber Operations Research Gym," *arXiv:2002.10667 [cs]*, 2 2020.
8. A. Molina-Markham, C. Miniter, B. Powell and A. Ridley, "Network Environment Design for Autonomous Cyberdefense," *CoRR,* vol. abs/2103.07583, 2021.
9. L. Li, R. Fayad and A. Taylor, "CyGIL: A Cyber Gym for Training Autonomous Agents over Emulated Network Systems," *CoRR,* vol. abs/2109.03331, 2021.
10. J. Schwartz and H. Kurniawatti, *NASim: Network Attack Simulator,* 2019.
11. Microsoft, *CyberBattleSim Project - Document and source code,* GitHub, 2021.
12. M. Standen, M. Lucas, D. Bowman, T. J. Richer, J. Kim and D. Marriott, "CybORG: A Gym for the Development of Autonomous Cyber Agents," *CoRR,* vol. abs/2108.09118, 2021
13. TTCP CAGE Challenges, GitHub - cage-challenge/cage-challenge-2: TTCP CAGE Challenge 2
14. G. Dulac-Arnold, D. Mankowitz and T. Hester, "Challenges of Real-World Reinforcement Learning," 2019.
15. A. Nair, M. Dalal, A. Gupta and S. Levine, "Accelerating Online Reinforcement Learning with Offline Datasets," *CoRR,* vol. abs/2006.09359, 2020.
16. X. B. Peng, M. Andrychowicz, W. Zaremba and P. Abbeel, "Sim-to-Real Transfer of Robotic Control with Dynamics Randomization," in *2018 IEEE International Conference on Robotics and Automation (ICRA)*, Brisbane, Australia, May, 2018
17. MITRE Corp, *MITRE ATT&CK knowledge base,* 2021.
18. VMWare Vsphere documentation, https://docs.vmware.com/en/VMware-vSphere/index.html
19. OpenAI, Gym Documentation, https://www.gymlibrary.dev, 2022
20. MITRE Corp., *CALDERA - Document and source code,* GitHub, 2021.
21. J. Farebrother, M. C. Machado and M. Bowling, "Generalization and Regularization in DQN," *CoRR,* vol. abs/1810.00123, 2018.
22. J. Schulman, F. Wolski, P. Dhariwal, A. Radford and O. Klimov, "Proximal Policy Optimization Algorithms," *CoRR,* vol. abs/1707.06347, 2017.
23. M. G. Bellemare, W. Dabney and R. Munos, "A distributional perspective on reinforcement learning," in *International Conference on Machine Learning*, 2017.